\documentclass[runningheads]{llncs}
\usepackage[T1]{fontenc}
\usepackage{amsmath}
\usepackage{amsfonts}
\usepackage{amssymb}
\usepackage{algorithm}
\usepackage{booktabs}
\usepackage{algpseudocode}
\usepackage{array}
\usepackage{tabularx}

\DeclareMathOperator{\EX}{\mathbb{E}}

\usepackage{graphicx}
\begin{document}
%

\title{Improving Out-of-Distribution Data Handling and Corruption Resistance via Modern Hopfield Networks}
%
%
\author{Saleh Sargolzaei\inst{1}\and
Luis Rueda\inst{1}}
\authorrunning{S. Sargolzaei and L. Rueda}
\titlerunning{Improving Corruption Resistance via Modern Hopfield Networks}
%
\institute{School of Computer Science\\
University of Windsor, Windsor, Canada \\
\email{\{sargolz,lrueda\}@uwindsor.ca}}
\maketitle              
\begin{abstract}
This study explores the potential of Modern Hopfield Networks (MHN) in improving the ability of computer vision models to handle out-of-distribution data. While current computer vision models can generalize to unseen samples from the same distribution, they are susceptible to minor perturbations such as blurring, which limits their effectiveness in real-world applications. We suggest integrating MHN into the baseline models to enhance their robustness. This integration can be implemented during the test time for any model and combined with any adversarial defense method. Our research shows that the proposed integration consistently improves model performance on the MNIST-C dataset, achieving a state-of-the-art increase of 13.84\% in average corruption accuracy, a 57.49\% decrease in mean Corruption Error (mCE), and a 60.61\% decrease in relative mCE compared to the baseline model. Additionally, we investigate the capability of MHN to converge to the original non-corrupted data. Notably, our method does not require test-time adaptation or augmentation with corruptions, underscoring its practical viability for real-world deployment. (Source code publicly available at: \url{https://github.com/salehsargolzaee/Hopfield-integrated-test})

\end{abstract}
\section{Introduction}
The effectiveness of modern computer vision algorithms relies heavily on the assumption that data is independent and identically distributed (i.i.d.). Consequently, challenges arise in adapting these algorithms to generalize to out-of-distribution data in real-world scenarios, where visual corruption caused by adverse weather, lighting variations, and other factors is prevalent \cite{foggyScene}. Studies have revealed a significant decrease in the generalization capability of models trained on clean data when exposed to these corruptions \cite{imageNetC,mnistC}.

In this study, we propose integrating a pre-trained Modern Hopfield Network (MHN) associative memory as an input module for the current models. Associative memory is capable of recovering the original input based on partial information \cite{standardHopfield}. We suggest that adding this module, pre-trained to remove Gaussian noise from clean data, would help recover the original data in case of various corruptions. The proposal has advantages when compared with successful methods in combating corruption, including test-time adaptation (TTA) \cite{2023tta,2021tta} and domain adaptation (DA) \cite{2020DA}.  

TTA requires updating the model weights during the test time to adapt to corrupted data in test data. This adaptation process can introduce new challenges caused by batch size \cite{ttaChallenge1} or temporary traits of test data \cite{2023tta,2018ddaSurvey}. However, our method does not require any test-time adaptation and can be trained offline and used during the test-time. 

DA involves adjusting models that have been trained on one set of data (the source - in this case, clean images) so that they can be used on another set for which only unlabeled samples are available (the target - in this case, the corrupted images) \cite{2020DA}. This process is most effective when source and target domain data are available simultaneously. However, our method does not require access to the target corruptions, making it more suitable for real-world applications. 

Notably, the proposed method can still be combined with the above-mentioned techniques. Our contributions can be summarized as follows:

\begin{itemize}
    \item We propose a general pre-training scheme with Modern Hopfield Networks to build generic extensions that enhance test-time robustness against corruptions for any baseline model trained on a clean dataset.
    \item We develop a test-time integration algorithm using the pre-trained extension and validate its effectiveness on the MNIST-C dataset.
   \item We demonstrate the superiority of modern Hopfield networks in tolerating various types of corruption, beyond just noise, by comparing our extension with a convolutional denoising autoencoder pre-trained using the same scheme.
 \item We provide insights into the robustness of our integration algorithm when incorporating non-effective modules. We show that the algorithm maintains baseline performance even with an ineffective convolutional denoising autoencoder.
    \item We demonstrate the superiority of our algorithm compared to other offline methods designed to handle unseen data (corruptions). We also show that our method is comparable to test-time adaptive (TTA) methods and can be combined with TTA or offline methods to enhance robustness further.
\end{itemize}

\section{Problem Statement}

We focus on a new approach to address the issue of encountering out-of-distribution and corrupted data during testing. Adapting individual models to different types of corruption and changes in distribution can be time-consuming and repetitive. Our goal is to develop a memory layer capable of swiftly retrieving clean or sufficiently clean data from corrupted inputs in real time. If successful, this layer can be integrated into any pre-trained classifier trained on clean data, thereby enhancing its robustness and adaptability.

Consider a classifier \( f(x) \) trained on a dataset \(\{(x_{i}, y_{i})\}_{i}\), where \((x_{i}, y_{i}) \sim \mathcal{D}\). Let \( C \) be a set of corruption functions, and let \(\mathbb{P}_{C}(c)\) represent the approximate frequency of corruption \( c \in C \) in the real world. The task of corruption robustness can be defined as follows \cite{imageNetC}:

\begin{equation} \label{equation1}
    \EX_{c \sim \mathbb{P}_{C}}\left[\mathbb{P}_{(x, y) \sim \mathcal{D}} \left(f\left( c(x)\right) = y\right)\right] \, ,
\end{equation}

\noindent where \(\mathbb{P}_{(x, y) \sim \mathcal{D}} \left(f\left( c(x)\right) = y\right)\) is the probability that the classifier \( f \) correctly classifies the corrupted input \( c(x) \). Our goal is to find an associative memory function \( h \) such that \((h(c(x_{i})), y_{i})\) is approximately distributed as \(\mathcal{D}\). We consider this goal achieved if:

\begin{equation} \label{equation2}
    \mathbb{P}_{(x, y) \sim \mathcal{D}} \left(f\left( h(c(x))\right) = y\right) \approx \mathbb{P}_{(x, y) \sim \mathcal{D}} \left(f\left(x\right) = y\right) \, ,
\end{equation}

\noindent where \(\mathbb{P}_{(x, y) \sim \mathcal{D}} \left(f\left(x\right) = y\right)\) is the probability that the classifier \( f \) correctly classifies the original input \( x \).

\section{Modern Hopfield Networks}

The classical Hopfield network was introduced as an associative memory model \cite{standardHopfield}. The model can be formalized as a system with $N$ binary neurons where activity of the neurons at time $t$ can be represented by a N-dimensional state vector $\boldsymbol{\sigma}^{(t)} = (\sigma_{i}^{(t)})_{i=1}^{N}$. In the original model, the states were considered to be binary, where $\sigma_{i}^{(t)} \in \left\{-1, +1\right\}$. In the classical Hopfield network, the update rule for each neuron is given by \cite{standardHopfield}:

\begin{equation} \label{equation3}
\sigma_{i}^{(t+1)} = Sign \left[ \sum_{j=1}^{N}T_{ij}\sigma_{j}^{(t)} \right] = Sign\left[\boldsymbol{T}\boldsymbol{\sigma}^{(t)}\right]_{i} \, ,
\end{equation}

\noindent where $\boldsymbol{T}$ is a symmetric real-valued connection matrix with zeros on the main diagonal, specifying the pairwise connection strength among neurons. It can be shown that this update rule may result in a monotonic decrease of the following energy function:

\begin{equation} \label{equation4}
E = -\sum_{i, j=1}^{N}\sigma_{i}T_{ij}\sigma_{j} = -\boldsymbol{\sigma}^{T}\boldsymbol{T}\boldsymbol{\sigma} \, .
\end{equation}

\noindent Therefore, by following the update rule in Equation (\ref{equation3}), the energy function may converge to a local minima. These local minima can be utilized to store memory (pattern), in such a way that by applying the update rule on a corrupted initialization of the memory, the original memory can be retrieved. The classic method for storing these memories involves encoding them in the weight matrix using the Hebb rule, which in its simplest form is:

\begin{equation} \label{equation5}
T_{ij} = \sum_{\mu=1}^{K}\xi_{i}^{\mu}\xi_{j}^{\mu} \, ,
\end{equation}

\noindent where the set of vectors $\{\boldsymbol{\xi}^{\mu}\}_{\mu = 1}^{K}$ represent $K$ patterns one wishes to store. In the classic Hopfield network, it has been shown that the maximal storage capacity in the case of random memories is in the order of $K \approx 0.14N$ \cite{abu1985information,mceliece1987capacity}. However, the network's storage capacity can be increased by introducing non-linear functions to the energy, which may result in higher than quadratic interactions between the neurons \cite{Demircigil_2017,krotov2016dense}. The following general form of energy function can characterize these Modern Hopfield Networks (MHN):

\begin{equation} \label{equation6}
E = -h\left(\sum_{\mu=1}^{K}F(\boldsymbol{\sigma}^{T}\boldsymbol{\xi}^{\mu})\right) \, ,
\end{equation}

\noindent where $F(\cdot)$ represents a rapidly growing smooth function and $h(\cdot)$ is a strictly monotonic and differentiable function that can preserve the stability and locations of local minima. Setting $F(x) = e^x$ has been proved to result in a theoretical capacity of $K \approx e^{\alpha N}, \quad \alpha < \frac{ln(2)}{2}$, which is exponential in the number of neurons $N$ \cite{Demircigil_2017}. We utilize the continuous state MHN introduced in \cite{ramsauer2021hopfield}. The energy function of this model is obtained by setting $h(x) = log(x)$ and $F(x) = e^{\beta x}$ in Equation (\ref{equation6}), where $\beta$ is a positive value. Due to the generalization to a continuous state vector, $\boldsymbol{\sigma} \in \Re^{N}$, regularization terms were added to ensure that the energy is bounded and the norm of the state vector remains finite. The proposed energy function was expressed as follows:

\begin{equation} \label{equation7}
E = - \beta^{-1}\log\left(\sum_{\mu=1}^{K}e^{\beta\boldsymbol{\sigma}^{T}\boldsymbol{\xi}^{\mu}}\right) + \frac{1}{2}\boldsymbol{\sigma}^{T}\boldsymbol{\sigma} + \beta^{-1}\log{K} + \frac{1}{2} M^2\, ,
\end{equation} 

\noindent where $M = \max_{\mu}\left\|\boldsymbol{\xi}^{\mu}\right\|$. The update rule for minimizing this energy is:

\begin{equation} \label{equation8}
\boldsymbol{\sigma}^{(t+1)} = \boldsymbol{X}\text{softmax}(\beta\boldsymbol{X}^{T}\boldsymbol{\sigma}^{(t)}) \, ,
\end{equation}

\noindent where $\boldsymbol{X} = (\boldsymbol{\xi}^1, \dots, \boldsymbol{\xi}^K)$ forms a matrix by stacking memory vectors as its columns. In the following sections, we use a particular version of this model.

\section{Methodology}

We propose a two-step approach to enhance the model's robustness against data corruption. First, we train a ``HopfieldPooling'' layer, a specialized variant of the continuous state Modern Hopfield Network (MHN) as introduced by Ramsauer et al. \cite{ramsauer2021hopfield}, on a denoising task. The primary objective is to develop an algorithm to integrate this trained HopfieldPooling module into an existing baseline model during the testing phase, thereby improving its resilience to various forms of data corruption. Both the baseline model and the HopfieldPooling layer are trained using clean data without access to future corrupted data during training.

\subsection{Denoising Task}

The denoising task serves as the preliminary step to achieve our main objective. Denoising tasks are widely used to learn robust data representations \cite{vincent2008extractingDenoise}. In this task, given an original input $\boldsymbol{x}$, we generate a corrupted version $\Bar{\boldsymbol{x}}$ by adding Gaussian noise with mean zero and standard deviation 0.5. Consequently, $\Bar{\boldsymbol{x}}$ can be modeled as a random variable following the distribution:

\begin{equation} \label{equation9}
\Bar{\boldsymbol{x}} | \boldsymbol{x} \sim \mathcal{N}(\boldsymbol{x},,0.5^{2}\boldsymbol{I}) \, .
\end{equation}

We train the HopfieldPooling layer to minimize the mean squared error (squared $L_2$-norm) between the original and denoised inputs, as defined by the following objective function:

\begin{equation} \label{equation10}
\min_{h} \frac{1}{m}\sum_{i=1}^{m}\left\|\boldsymbol{x}^{i} - h(\Bar{\boldsymbol{x}}^{i})\right\|_{2}^{2}\, ,
\end{equation}

\noindent where $\{\boldsymbol{x}^{i}\}_{i = 1}^{m}$ represents the set of training vectors and $h$ denotes the HopfieldPooling layer.

\subsection{Integration Algorithm}

Once the HopfieldPooling layer is trained on the denoising task, we propose Algorithm \ref{alg:algo1} to integrate this module into an existing baseline model during the testing phase. The algorithm aims to process both corrupted data and the corrected version through the HopfieldPooling layer. It then makes predictions based on the input that generates more confidence for the most confident class.
This integration aims to enhance the model's ability to handle corrupted data effectively, leveraging the learned memory patterns from the HopfieldPooling layer to correct or mitigate the effects of corruptions encountered during testing.

\begin{algorithm}
\caption{Test Model with Hopfield Integration.}
\small
\begin{algorithmic}[1]
    \State \textbf{Input:} $\mathcal{D}_{test} = \{\boldsymbol{x}^i\}_{i=1}^M$ \Comment{Test dataset}
    \State \textbf{Output:} $\mathcal{P} = \{p_i\}_{i=1}^M$ \Comment{Predictions for test data}
    \State $\mathcal{P} \gets \emptyset$ \Comment{Initialize the set of predictions}
    
    \ForAll{$\boldsymbol{x}^i \in \mathcal{D}_{test}$}
        \State $\mathbf{o}_f \gets f(\boldsymbol{x}^i)$ \Comment{Base model output (vector of class probabilities)}
        \State $\mathbf{o}_h \gets f(h(\boldsymbol{x}^i))$ \Comment{Hopfield integrated output (vector of class probabilities)}
        
        \State $(\text{max\_prob}_f, \text{pred\_class}_f) \gets (\max_j(\mathbf{o}_f[j]), \arg\max_j(\mathbf{o}_f[j]))$
        \State $(\text{max\_prob}_h, \text{pred\_class}_h) \gets (\max_j(\mathbf{o}_h[j]), \arg\max_j(\mathbf{o}_h[j]))$
        
        \If{$\text{max\_prob}_f > \text{max\_prob}_h$}
            \State $p_i \gets \text{pred\_class}_f$ \Comment{Choose base model prediction}
        \Else
            \State $p_i \gets \text{pred\_class}_h$ \Comment{Choose Hopfield integrated prediction}
        \EndIf
        
        \State $\mathcal{P} \gets \mathcal{P} \cup \{p_i\}$ \Comment{Update test results with final prediction}
    \EndFor
    
    \State \Return $\mathcal{P}$ \Comment{Return all predictions}
\end{algorithmic}
\normalsize
\label{alg:algo1}
\end{algorithm}

\section{Experiments}

\subsection{Experimental Setup}
We conducted our experiments in the Google Colab environment using an NVIDIA Tesla T4 GPU with 15360 MiB of memory. We performed the training phase using Python 3.10.12 and PyTorch 2.3.0+cu121. 

\subsubsection{Dataset:}
For training purposes, we used $60,000$ clear training samples from the MNIST dataset \cite{lecun1998mnist}. To test the proposed method, we utilized the MNIST-C dataset, a benchmark designed to evaluate the robustness of computer vision models \cite{mnistC}. This dataset was created by applying 15 types of corruptions to the $10,000$ test images from the clean MNIST dataset, resulting in a total of $150,000$ corrupted images. The corruptions include shot noise, impulse noise, glass blur, motion blur, shear, scale, rotate, brightness, translate, stripe, fog, spatter, dotted line, zigzag, and canny edges.

\subsubsection{Implementation Details:} 
To ensure repeatability, we trained and used the default convolutional neural network provided by the official PyTorch repository \cite{pytorchMNIST} without any modifications as the base model. This model definition is also employed in the benchmark paper of the MNIST-C dataset \cite{mnistC}. For training the HopfieldPooling layer, we used the hyperparameters shown in Table \ref{tab1}, which are inspired by the examples provided in the original publication's repository \cite{hopfieldLayers}.
For optimization, we adopted the AdamW optimizer introduced in \cite{loshchilov2019decoupled}, with the following default parameters:  $\alpha = 0.001$, $\beta_{1} = 0.9$, $\beta_{2} = 0.999$, $\epsilon = 10^{-8}$, and $\lambda = 0.01$. Our training process involves 20 epochs and a batch size of 20.

\begin{table}
\caption{Hyperparameters used for training the HopfieldPooling layer.}\label{tab1}
\begin{tabularx}{\columnwidth}{l c X}%
\toprule
Hyperparameter &  Value &  Description\\
\midrule
input\_size &  {784} & Dimension of the input vector $\Bar{\boldsymbol{x}}$ in Equation (\ref{equation9})\\ 
hidden\_size &  {8} & Dimension of the association space $N$\\
num\_heads & {8} & Number of parallel Hopfield heads\\
update\_steps\_max & {5} & Number of updates for each Hofield head in each epoch \\
scaling & {0.25} &  $\beta$ parameter in Equation (\ref{equation8})\\
\bottomrule
\end{tabularx}
\end{table}

\subsubsection{Evaluation Metrics:}
To evaluate the robustness gained by integrating the HopfieldPooling module into any baseline model, we use the mean Corruption Error (mCE) and relative mCE metrics, as established in benchmark publications \cite{imageNetC,mnistC}. Given a corruption function \( c \in C \), a classifier \( g \), and a baseline classifier \( f \), we denote their error rates on \( c \) as  $E_{c}^{g}$ and $E_{c}^{f}$, respectively.

To account for varying corruption difficulties, we calculate the corruption error of \( g \) on \( c \) ($CE_{c}^{g}$) by normalizing its error rate with the baseline error rate:

\begin{equation} \label{equation11}
    CE_{c}^{g} = \frac{E_{c}^{g}}{E_{c}^{f}}
\end{equation}

We also assess degradation relative to clean data (identity corruption \( i \)) by calculating the relative corruption error:

\begin{equation} \label{equation12}
   \text{relative } CE_{c}^{g} = \frac{E_{c}^{g} - E_{i}^{g}}{E_{c}^{f} - E_{i}^{f}}
\end{equation}

From the two metrics defined above, we compute the mean values across all corruptions, resulting in mean CE (mCE) and relative mCE. Consequently, the mCE and relative mCE of the baseline model \( f \) are expected to be one (or 100\%) due to these calculations. To quantify robustness improvements, we measure the reduction in mCE from 100\%. Additionally, we calculate the gain in average corrupted accuracy.

\subsection{Integration Results}

Table \ref{tab2} shows the evaluation metrics for the baseline model and Hopfield integration. The relative mCE and mCE are both 100\% for the baseline model. With Hopfield integration, these values significantly decrease to 39.39\% and 42.51\%, respectively, demonstrating substantial improvements in robustness with reductions of 60.61 and 57.49 percentage points. Additionally, the average corruption accuracy increases from 75.92\% for the baseline model to 89.76\% with Hopfield integration, marking an improvement of 13.84 percentage points.

\begin{table}[h!]
\caption{Comparison of different evaluation metrics with and without Hopfield integration. Symbols $\downarrow$ and  $\uparrow$ denote the desired direction of change for each metric.}\label{tab2}
\begin{tabularx}{\columnwidth}{l X X X}%
\toprule
Metric & Baseline & Hopfield-integrated & Improvement \\
\midrule
Relative mCE (\%) $\downarrow$ & 100 & \textbf{39.39} &  60.61\\
mCE (\%) $\downarrow$ & 100 & \textbf{42.51} & 57.49\\
Average Corruption Accuracy (\%) $\uparrow$ & 75.92 & \textbf{89.76} & 13.84 \\
\bottomrule
\end{tabularx}
\end{table}

To better understand the effect of the HopfieldPooling layer on different corruptions, we compared the corrupted accuracy for each type. Table \ref{tab3} displays these results. It can be observed that the corruption accuracy of the Hopfield-integrated model generally surpasses that of the baseline model. Specifically, the integration yields significant accuracy improvements of 82.65\%, 50.13\%, 43.42\%, 18.41\%, and 14.32\% for fog, glass\_blur, motion\_blur, impulse\_noise, and brightness, respectively. Conversely, minor reductions or slight improvements in accuracy are observed for affine transformations, with changes of -3.06\%, -0.94\%, -0.68\%, and 0.11\% for translate, rotate, scale, and shear, respectively. These results highlight that the improvements are substantially more significant and robust, enhancing the baseline model's performance under severe levels of corruption. 

\begin{table}
\caption{Classification accuracy (\%) of the baseline and Hopfield-integrated models on different types of corruptions.}\label{tab3}
\begin{tabularx}{\columnwidth}{l X X}%

\toprule
Corruption & Baseline & Hopfield-integrated  \\
\midrule
identity (no corruption) & \textbf{99.04} & 98.87 \\
brightness & 82.66 & \textbf{96.98} \\
canny\_edges & 77.22 & \textbf{77.34} \\
dotted\_line & \textbf{98.19} & 98.01 \\
fog & 14.35 & \textbf{97.00} \\
glass\_blur & 39.66 & \textbf{89.79} \\
impulse\_noise & 77.49 & \textbf{95.90} \\
motion\_blur & 47.40 & \textbf{90.82} \\
rotate & \textbf{89.89} & 88.95 \\
scale & \textbf{89.10} & 88.42 \\
shear & 95.45 & \textbf{95.56} \\
shot\_noise & 96.40 & \textbf{97.71} \\
spatter & 97.78 & \textbf{97.96} \\
stripe & 95.40 & 95.40 \\
translate & \textbf{47.42} & 44.36 \\
zigzag & 90.40 & \textbf{92.26} \\
\midrule
Average Corruption Accuracy & 75.92 & \textbf{89.76} \\
\bottomrule
\end{tabularx}
\end{table}

Our integration algorithm makes the final decision based on both the corrupted input and the input provided by the HopfieldPooling layer. We investigated how often the HopfieldPooling layer was used for each corruption and how it affected accuracy. The results are shown in Table \ref{tab4}. The Pearson correlation coefficient was found to be $r = 0.637$ with a $p$-value of $0.008$. The coefficient suggests a moderate positive correlation between the usage of the HopfieldPooling layer and accuracy improvement. The $p$-value is well below the commonly accepted significance threshold of $0.05$, indicating that this correlation is statistically significant. This suggests that higher usage of the HopfieldPooling layer is associated with greater improvements in accuracy. 

Table \ref{tab4} also illustrates that the most significant improvements are obtained by utilizing the HopfieldPooling layer for almost all the decisions. For instance, in the case of fog and motion\_blur, the proposed integration utilized the HopfieldPooling layer for 99.94\% and 99.43\% of the decisions, respectively. Also, it turns out that, even in the case of minor or no improvement in accuracy, as in shot\_noise or shear, the baseline model utilized the HopfieldPooling layer for its final decisions. This behavior suggests that the Hopfield module also improves the final probabilities, and hence, the model's confidence in selecting the correct class.

\begin{table}[ht]
\centering
\caption{HopfieldPooling usage and increase in accuracy for different corruptions.}\label{tab4}
\begin{tabularx}{\columnwidth}{l c c}%
\toprule
Corruption & HopfieldPooling Usage (\%) & Increase in Accuracy (\%) \\
\midrule
identity (no corruption) & 20.47 & -0.17\\
brightness & 99.84 & 14.32\\
canny\_edges & 0.65 & 0.12\\
dotted\_line & 36.68 & -0.18\\
fog & 99.94 & 82.65\\
glass\_blur & 97.69 & 50.13\\
impulse\_noise & 86.36 & 18.41\\
motion\_blur & 99.43 & 43.42\\
rotate & 72.68 & -0.94\\
scale & 49.36 & -0.68\\
shear & 74.56 & 0.11\\
shot\_noise & 82.81 & 1.31\\
spatter & 50.55 & 0.18\\
stripe & 0.00 & 0.00\\
translate & 39.15 & -3.06\\
zigzag & 33.15 & 1.86\\
\midrule
\multicolumn{1}{r}{} & Pearson Correlation (r) & p-value \\
\cmidrule{2-3}
\multicolumn{1}{r}{} & 0.637 & 0.008 \\
\bottomrule
\end{tabularx}
\end{table}

\subsection{Ablation Study} 
To assess the potential of achieving similar results by integrating a different pre-trained denoising model, we replaced the HopfieldPooling layer with a stacked Convolutional Denoising Autoencoder (CDAE) \cite{masci2011stacked,vincent2010stacked}. This ablation study is vital for determining whether the improvements observed are specifically attributed to the embedded memories of the Hopfield associative memory or if similar results can be reproduced using alternative techniques, which ultimately would lead to constructing a robust latent representation of input data. 

\subsubsection{Convolutional Autoencoder Implementation:} Tables \ref{tab5} and \ref{tab6} provide the layers of the encoder and decoder parts of the convolutional autoencoder, respectively. After each 2D convolution and transposed convolution operation, a ReLU activation function is used, which is defined as $ReLU(x) = max(0, x)$. Since the original input values are between 0 and 1, the final output is passed through a Sigmoid activation function, which reduces the output logits to values between 0 and 1, and which is defined as $Sigmoid(x) = \frac{1}{1+e^{-x}}$. We keep the experimental setup constant (cf. Section 5.1).  

\begin{table}
\caption{Architecture of the encoder module.}\label{tab5}
\begin{tabularx}{\columnwidth}{l l c c c c c}%
\toprule
Layer & Operation & Number of Kernels & Kernel Size & Stride & Padding & Output Shape \\
\midrule
1&2D convolution & 32 & (3, 3) & 1 & 1 & (28, 28, 32)\\
2&2D max pooling  & - & (2, 2) & 2 & 0 & (14, 14, 32)\\
3&2D convolution  & 16 & (3, 3) & 1 & 1 & (14, 14, 16)\\
4&2D max pooling  & - & (2, 2) & 2 & 0 & (7, 7, 16)\\
5&2D convolution   & 8 & (3, 3) & 1 & 1 & (7, 7, 8)\\
6&2D max pooling  & - & (2, 2) & 2 & 0 & (3, 3, 8)\\
\bottomrule
\end{tabularx}
\end{table}

\begin{table}
\caption{Architecture of the decoder module.}\label{tab6}
\begin{tabularx}{\columnwidth}{l X c c c c c}%
\toprule
Layer & Operation & Number of Kernels & Kernel Size & stride & padding & Output Shape \\
\midrule
1  & 2D transposed convolution & 8 & (3, 3) & 2 & 0 & (7, 7, 8)\\
2  & 2D transposed convolution  & 16 & (2, 2) & 2 & 0 & (14, 14, 16)\\
3  & 2D transposed convolution  & 32 & (2, 2) & 2 & 0 & (28, 28, 32)\\
4  & 2D convolution   & 1 & (3, 3) & 1 & 1 & (28, 28, 1)\\
\bottomrule
\end{tabularx}
\end{table}

\subsubsection{Comparison on the Denoising Task:} 
\begin{figure}
\includegraphics[width=\textwidth]{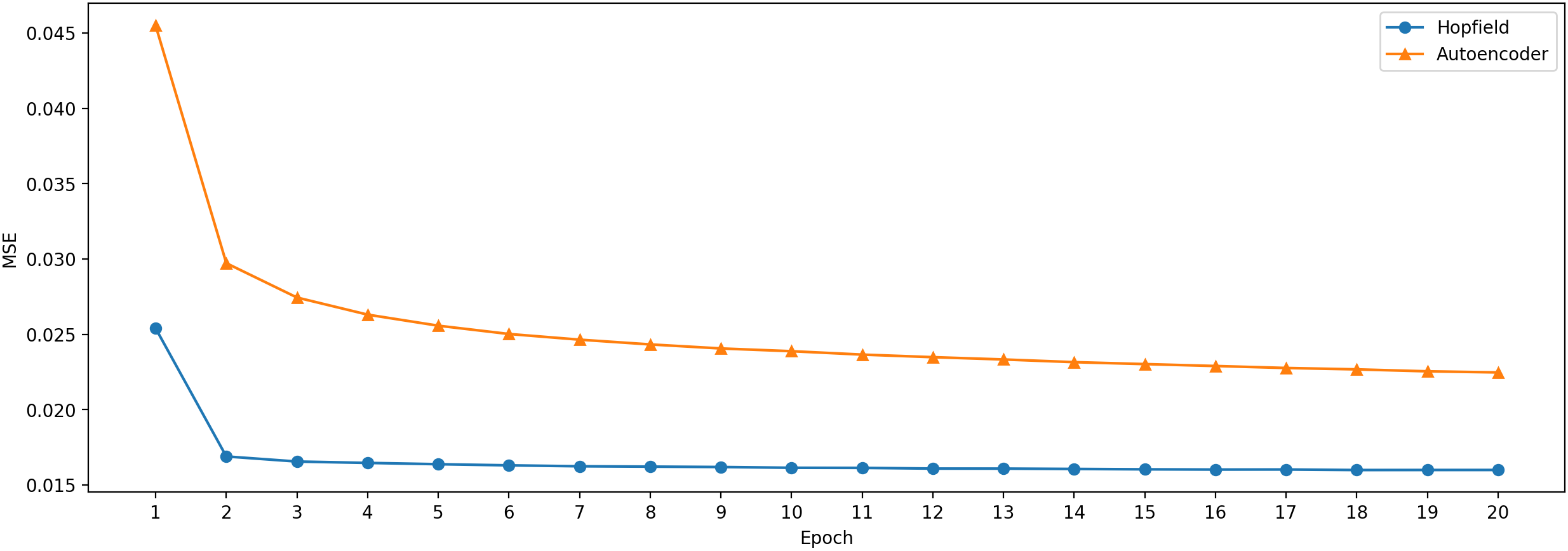}
\caption{Comparison of the mean squared error of different models per epoch during training on the denoising task.} \label{denoising_comp}
\end{figure}

Fig. \ref{denoising_comp} illustrates a comparison of the mean squared error (MSE) per epoch on the preliminary denoising task (cf. Section 4.1) for the HopfieldPooling layer and CDAE. We observe that HopfieldPooling exhibits significant superiority in terms of MSE. HopfieldPooling reaches a training error of 0.016 at the second training epoch, while CDAE plateaus near 0.022 even after 20 training epochs. 

\subsubsection{Comparison on the Integration Algorithm:} Fig. \ref{metric_comp} illustrates the robustness metrics across three conditions: baseline model, CDAE integration, and HopfieldPooling integration. The metrics indicate that CDAE integration offers minimal improvements. Furthermore, Pearson correlation analysis between the number of decisions based on CDAE outputs and corrupted accuracy improvements yielded a correlation coefficient (r) of 0.427 with a $p$-value of 0.099. Since the $p$-value exceeds the commonly accepted significance threshold of 0.05, this correlation is not statistically significant. Consequently, we cannot assert a meaningful linear relationship between decisions based on CDAE output and improvements in corrupted accuracy.

\subsubsection{Comparison of Corruption Removal:} 
\begin{figure}
\includegraphics[width=\textwidth]{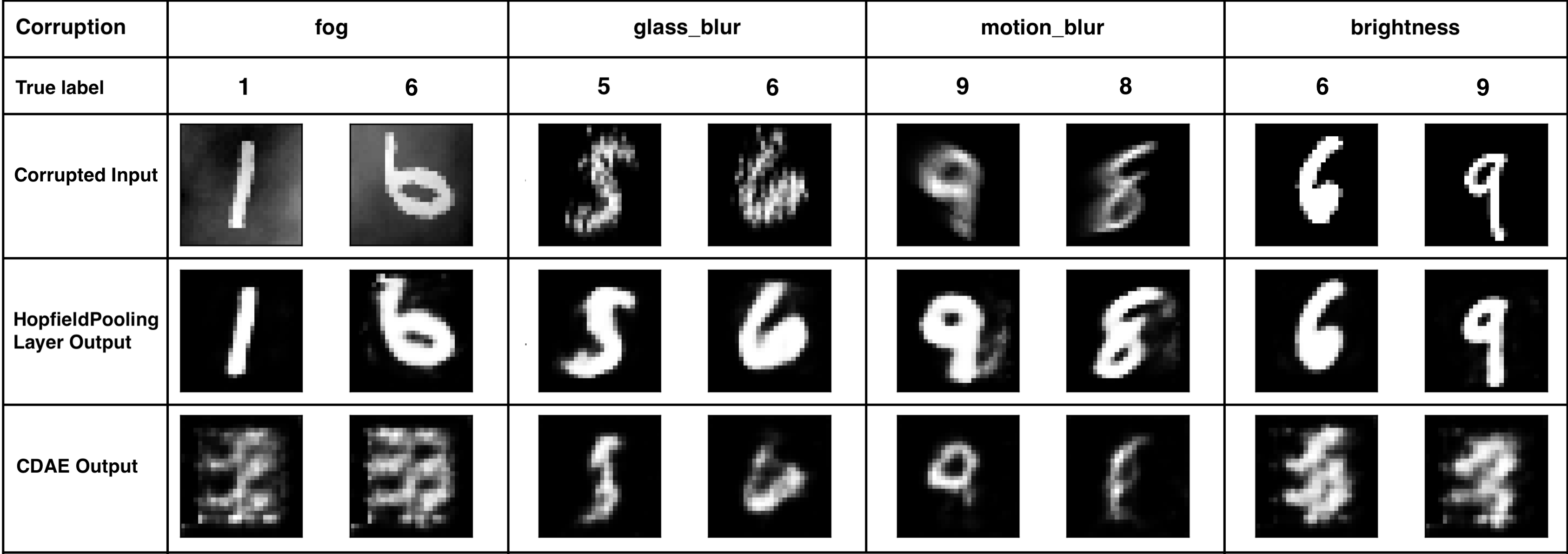}
\caption{The output of the HopfieldPooling layer and CDAE for corruptions on which notable improvement in robustness metrics is achieved with HopfieldPooling.} \label{best_improve}
\end{figure}

\begin{figure}
\includegraphics[width=\textwidth]{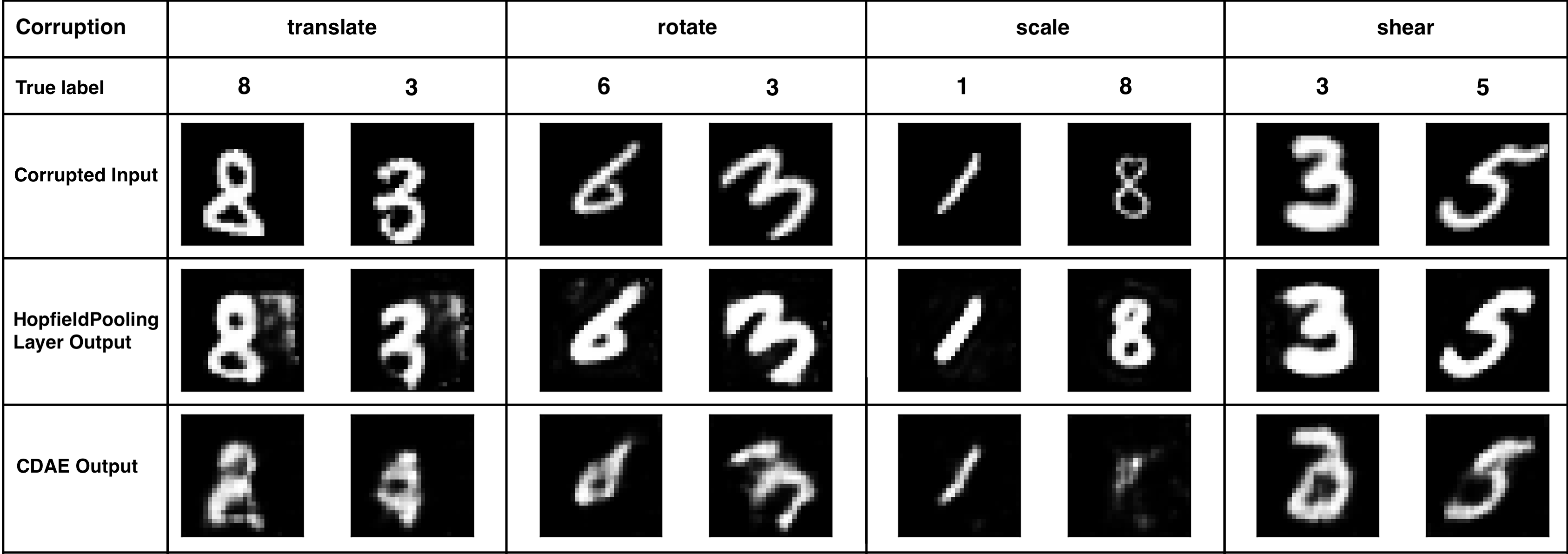}
\caption{The output of the HopfieldPooling layer and CDAE for corruptions made with affine transformations.} \label{affine}
\end{figure}

To investigate the challenges of CDAE in enhancing robustness compared to the HopfieldPooling layer, we analyzed the output of each pre-trained network when subjected to corrupted inputs. We present the outputs for both types of corruption where HopfieldPooling exhibited significant accuracy improvements (Fig. \ref{best_improve}) and the affine transformations where the metrics showed limited improvements (Fig. \ref{affine}).

\begin{figure}
\includegraphics[width=\textwidth]{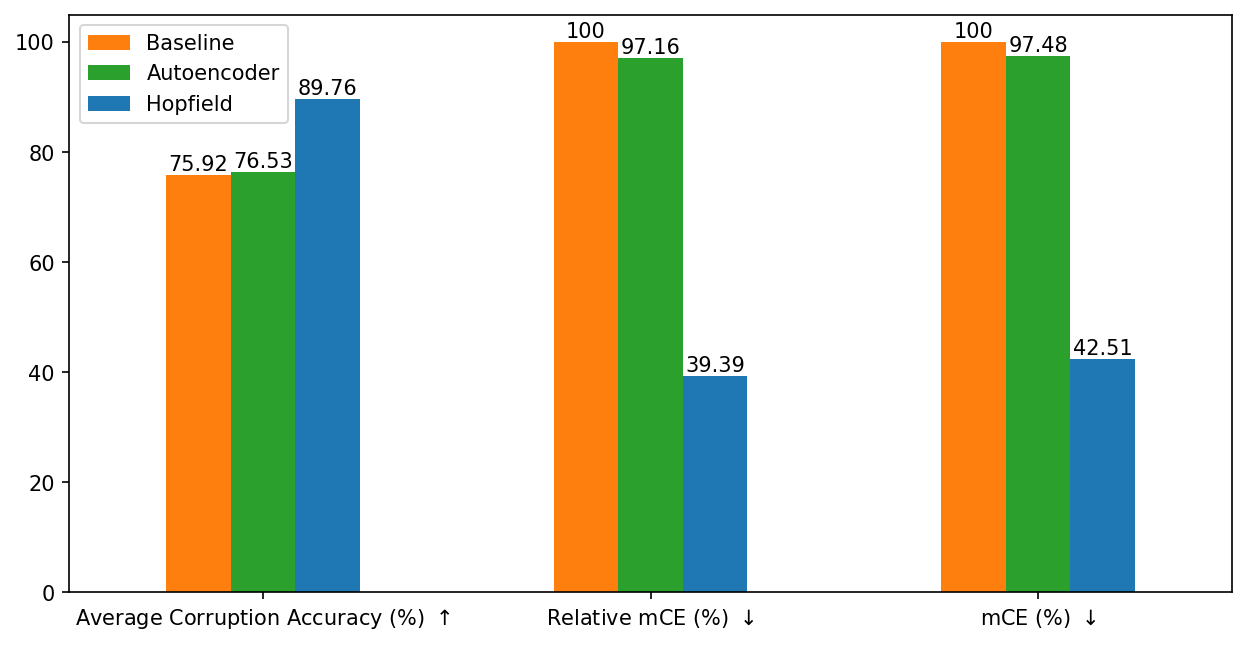}
\caption{Comparison of baseline, autoencoder-integrated, and Hopfield-integrated models on different robustness metrics during test time. Symbols $\downarrow$ and  $\uparrow$ denote each metric's desired direction of change.} \label{metric_comp}
\end{figure}

The illustrations demonstrate that CDAE not only fails to remove corruption but also disrupts the digit pattern entirely. In some cases, these disruptions make classification exremely difficult, even for humans. This failure could be due to the inability of the latent representation to generalize beyond Gaussian noise corruptions. Yet, these results underscore the robustness of our proposed integration algorithm to ineffective modules. As previously shown in Fig. \ref{metric_comp}, the integration algorithm still caused slight improvements by adding CDAE, suggesting that the algorithm mainly decides based on the best input. 

Conversely, the HopfieldPooling layer effectively mitigates most corruption. Notably, in the case of affine transformations shown in Fig. \ref{affine}, the model successfully reconstructs the digits despite not being trained on any affine transformations. These results suggest that the integration algorithm's inability to enhance robustness in affine transformations may be attributed to the base model's limitations. Previous studies have also indicated that convolutional neural networks (CNNs) are vulnerable to simple transformations such as translation and rotation \cite{engstrom2017rotation}. Nevertheless, the HopfieldPooling layer effectively generalizes to these transformations and reconstructs the correct digit. Additionally, it is observed that the digits become bolder, which may explain why the module increases the model's confidence in many cases.

\section{Discussion and Future Directions}

In this section, we situate our method within the current state-of-the-art research on robustness to data corruption and outline potential future research opportunities by integrating these methods. Table \ref{tab7} presents accuracy improvements achieved by training state-of-the-art computer vision models with various data augmentation techniques. Our method demonstrates superior performance across these methods. Additionally, we propose the HopfieldPooling layer as a versatile extension applicable to any baseline model. These augmentation techniques can be leveraged in future work to enhance the robustness of baseline models or to refine the training procedures for the Hopfield module. Furthermore, adversarial training methods could be employed in the training phase of the HopfieldPooling layer to develop a more generalized module resilient to a broader spectrum of corruptions \cite{stutz2020confidence}.

\begin{table}
\caption{Comparison of accuracy improvements (\%) using data augmentation methods and our method.}\label{tab7}
\begin{tabularx}{\columnwidth}{X c}%
\toprule
Method & Corrupted Accuracy Improvement $\uparrow$ \\ 
\midrule
Augment training data with 31 corruptions \cite{mnistC} & 6.39\\
Tuned training with additive Gaussian and Speckle noise \cite{rusak2020increasing} & 5.5\\
Augment training data with $\alpha-stable$ noise \cite{yuan2023robustnessenhancementneuralnetworks} & 8.85\\
Augment training data using multi-scale random convolutions \cite{xu2020robust} & 3.42 \\
\midrule
Hopfield integration method (ours) & \textbf{13.84} \\
\bottomrule
\end{tabularx}
\end{table}

Another promising direction is to incorporate our integration module into test-time adaptation (TTA) methods. Our method has already outperformed some of the leading TTA models, such as LAME \cite{boudiaf2022parameter}. For instance, Gong et al. \cite{gong2023noterobustcontinualtesttime} reported a classification error of 11.8 on the MNIST-C dataset using the LAME method, whereas our method achieved a classification error of 10.24 without any adaptation. Despite this, their TTA method (NOTE) reduced the classification error to 7.1. We propose further investigations into applying different TTA methods to adapt the general HopfieldPooling layer during test time. The adapted layer can then be used for more than one baseline model.

Finally, continuous-state modern Hopfield networks \cite{ramsauer2021hopfield} have gained significant attention in recent years. Our method exemplifies their application, yet numerous possibilities remain for future research. These include testing on color images, comparing various architectures and energy functions, and exploring the impact of hyperparameters on the Hopfield module.

\section{Conclusion}
Our study tackles the challenge of enhancing the reliability of computer vision models, particularly under test-time corruption. We introduce a universal integration algorithm that leverages a pre-trained modern Hopfield network on a clean dataset to significantly boost the performance of any baseline model. Our approach demonstrates comparable results to methods relying on extensive data augmentation or test-time adaptation, as evidenced by our experiments on the MNIST-C dataset. Moreover, our method's versatility allows for seamless integration with other techniques. The critical role of the Hopfield network was highlighted by our comparison with a convolutional denoising autoencoder, which did not yield significant improvements, further validating the effectiveness of our proposed approach. 

\subsubsection{Acknowledgements} This research work has been partially supported by the Natural Sciences and Engineering Research Council of Canada, NSERC, and the Vector Institute for Artificial Intelligence. 

%
%
%
\bibliographystyle{splncs04}
\bibliography{main}
\end{document}